\documentclass[journal]{IEEEtran}
\ifCLASSINFOpdf
\else
\fi
\usepackage[utf8]{inputenc}
\usepackage[english]{babel}
\usepackage{xcolor,colortbl}
\usepackage{epsfig}
\usepackage{graphicx}
\usepackage{amsmath}
\usepackage{amssymb}
\usepackage{tabularx}
\usepackage{algorithmic}
\usepackage{algorithm}
\usepackage{subcaption}
\usepackage{textcomp}
\usepackage{placeins}
\usepackage{epstopdf}
\usepackage{multirow}
\usepackage{siunitx}
\usepackage{makecell,multirow}
\usepackage{subcaption}
\usepackage[utf8]{inputenc}
\usepackage{hyperref}
\usepackage{caption}
\usepackage{fancyhdr}

\begin{document}
	%
	\title{Adversarial Network with Multiple Classifiers for Open Set Domain Adaptation}
	%
	%
	
	\author{Tasfia~Shermin,
		Guojun~Lu,~\IEEEmembership{Senior Member,~IEEE,}
		Shyh Wei~Teng,
		Manzur~Murshed,~\IEEEmembership{Senior Member,~IEEE,}
		and~Ferdous~Sohel,~\IEEEmembership{Senior Member,~IEEE}
		\thanks{Tasfia Shermin, Guojun Lu, Shyh Wei Teng, and Manzur Murshed are with the School of Science, Engineering and Information Technology, Federation University, Churchill-3842, Australia (email: \{t.shermin, guojun.lu, shyh.wei.teng, manzur.murshed\}@federation.edu.au)}
		\thanks{Ferdous Sohel is with Murdoch University, WA-6150, Australia (email: f.sohel@murdoch.edu.au)}}
	\maketitle
	
	\begin{abstract}
		Domain adaptation aims to transfer knowledge from a domain with adequate labeled samples to a domain with scarce labeled samples. Prior research has introduced various open set domain adaptation settings in the literature to extend the applications of domain adaptation methods in real-world scenarios. This paper focuses on the type of open set domain adaptation setting where the target domain has both private (`unknown classes') label space and the shared (`known classes') label space. However, the source domain only has the `known classes' label space. Prevalent distribution-matching domain adaptation methods are inadequate in such a setting that demands adaptation from a smaller source domain to a larger and diverse target domain with more classes. For addressing this specific open set domain adaptation setting, prior research introduces a domain adversarial model that uses a fixed threshold for distinguishing known from unknown target samples and lacks at handling negative transfers. We extend their adversarial model and propose a novel adversarial domain adaptation model with multiple auxiliary classifiers. The proposed multi-classifier structure introduces a weighting module that evaluates distinctive domain characteristics for assigning the target samples with weights which are more representative to whether they are likely to belong to the known and unknown classes to encourage positive transfers during adversarial training and simultaneously reduces the domain gap between the shared classes of the source and target domains. A thorough experimental investigation shows that our proposed method outperforms existing domain adaptation methods on a number of domain adaptation datasets.
		
	\end{abstract}
	
	%
	\IEEEpeerreviewmaketitle
	
	\section{Introduction}	
	\IEEEPARstart{D}{eep} learning models for computer vision tasks usually require a massive amount of labeled data entailing highly laborious work for annotating data \cite{szegedy2015going, he2016deep, szegedy2017inception, tshermin2019}. An alternative is to use labeled data from a related (source) domain to boost the performance of the model in a target domain. However, as the source and target data may have domain gaps such as different illumination set-ups, and perspectives, synthesized data by using different variants of sensors, the performance of this approach may suffer. Existing domain adaptation (DA) methods aim to decrease the above-mentioned domain divergences either by using distribution matching methods \cite{bousmalis2017unsupervised,  ganin2015unsupervised, li2018heterogeneous, li2019locality} or by transforming samples from one domain to another through generative models \cite{8656504, bousmalis2017unsupervised, tzeng2017adversarial, liu2017unsupervised, taigman2016unsupervised, saito2018maximum, li2018transfer}.
	
\begin{figure*}[!ht]
		\centering
		\begin{subfigure}[b]{.47\textwidth}
			\centering
			\includegraphics[width=.8\textwidth]{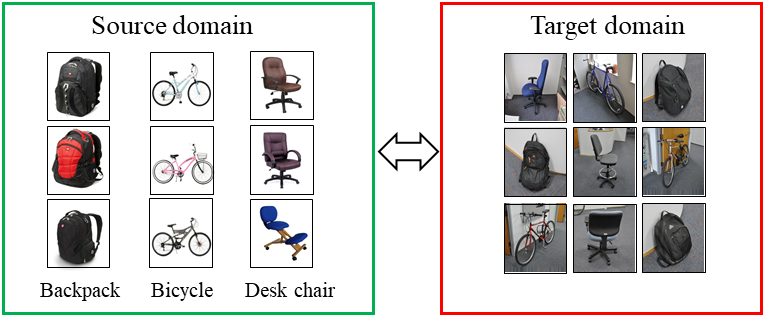}
			\caption[]%
			{{\small Closed set domain adaptation setting}}    
			\label{1a}
		\end{subfigure}
		\hfill
		\begin{subfigure}[b]{.47\textwidth}  
			\centering 
			\includegraphics[width=.8\textwidth]{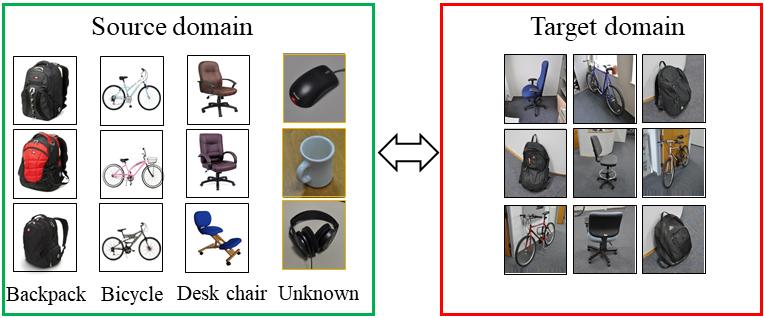}
			\caption[]%
			{{\small Partial domain adaptation setting}}    
			\label{1b}
		\end{subfigure}
		\vskip\baselineskip
		\begin{subfigure}[b]{.47\textwidth}   
			\centering 
			\includegraphics[width=.8\textwidth]{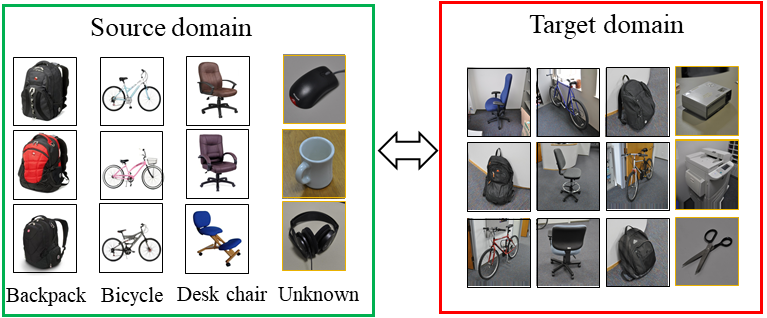}
			\caption[]%
			{{\small Open set domain adaptation setting \cite{panareda2017open}}  }  
			\label{1c}
		\end{subfigure}
		\hfill
		\begin{subfigure}[b]{.47\textwidth}   
			\centering 
			\includegraphics[width=.8\textwidth]{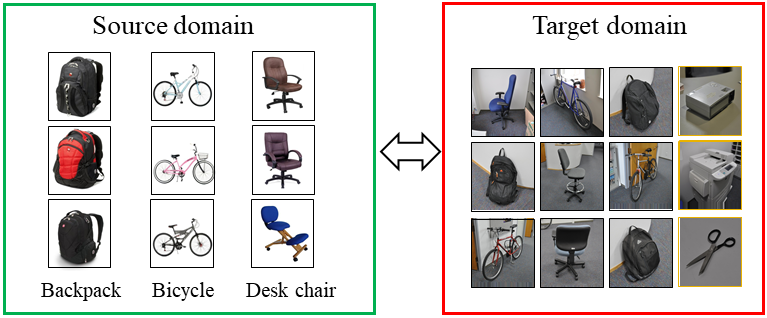}
			\caption[]%
			{{\small Open set domain adaptation by back-propagation setting \cite{saito2018open}} }   
			\label{1d}
		\end{subfigure}
		\caption[ ]
		{(a) The closed set domain adaptation setting assumes that both source and target domains consist of images only of the same set of classes. (b) Partial domain adaptation setting assumes that the source domain label space (classes) is the superspace of the target domain label space. This adaptation setting includes images of unknown classes in the source domain only. (c) Open set domain adaptation setting proposed by Busto et al. \cite{panareda2017open} requires images of unknown classes in both source and target domains, i.e., both source and target domains contain images that do not belong to the label space of interest. (d) Open set domain adaptation setting introduced by Saito et al. \cite{saito2018open} requires images of unknown classes only in the target domain besides the classes of interest. We focus on the domain adaptation setting illustrated in (d).  
		} 
		\label{1}
	\end{figure*}

	Generally, it is assumed that label sets across the source and target domains are identical (closed set domain adaptation \cite{tzeng2014deep, long2016unsupervised, haeusser2017associative}), as shown in Fig. \ref{1a}. However, such a simplified setting only has limited real-world applications. Open set domain adaptation \cite{saito2018open, panareda2017open}, and partial domain adaptation \cite{cao2019learning, cao2018partial, zhang2018importance} methods have been proposed to ease the closed set domain adaptation assumption. Open set and partial domain adaptation settings assume source or target domain private label sets besides the identical (shared) label sets. The domain adaptation models-based on these domain adaptation settings are required to recognize the samples of the target domain private label sets as `unknown' class and the samples belonging to shared label sets as known classes.
	
	As illustrated in Fig. \ref{1b}, partial domain adaptation \cite{cao2019learning, cao2018partial, zhang2018importance} setting assumes that the source domain label space is a superspace of the target domain label space. Open set domain adaptation setting proposed by Busto et al. \cite{panareda2017open} requires images from unknown classes both in the source and target domain besides the shared known classes. 
	However, this is not a cost-effective setting for open set domain adaptation as it requires a collection of a large number of unknown source samples with no prior knowledge about target labels. During training, Busto et al. \cite{panareda2017open} bound the domain adaptation model to align unknown classes of the target domain towards unknown classes of the source domain. This may enforce a firm boundary for the unknown classes. During testing, samples from unknown classes other than the trained unknown classes may confuse the model.
	
	Open set domain adaptation by back-propagation (OSBP) \cite{saito2018open} setting takes another step towards the practical domain adaptation scenario by removing unknown classes from the source domain such that the source label space is a subset of the target label space (Fig. \ref{1d}). This means that the OSBP domain adaptation setting has samples from unknown classes only in the target domain and encourages the domain adaptation model to learn to detect an unknown target sample as unknown when it does not belong to the known classes. Thus, this setting assists in training the domain adaptation model with a broader unknown class boundary and detect unknown samples during testing better than the previous setting \cite{panareda2017open}. This paper focuses on improving the performance of domain adaptation model for the OSBP domain adaptation setting, which seems to be realistic and challenging as the source domain has less number of classes compared to the target domain.
	
	
	Prevalent distribution matching domain adaptation methods \cite{bousmalis2017unsupervised, ganin2015unsupervised} cannot be applied to the OSBP domain adaptation setting as the absence of unknown samples in the source domain does not allow the unknown samples of the target domain to be aligned. Saito et al. \cite{saito2018open} have proposed a generative model to address the OSBP domain adaptation setting where they enable the classifier to draw a rough boundary between the source and target samples (i.e., initially all the target samples will be classified as `unknown') and the generator has to separate the target samples into known and unknown classes adversarially. The adversarial learning between the generator and the classifier depends on the pseudo decision of the classifier. However, their proposed method does not explore any underlying domain discriminative information to assess the pseudo decision of the classifier before the adversarial training. Also, they set an empirical fixed threshold for generator-classifier adversarial training to differentiate known target samples from unknown ones. We argue that this concept of relying only on the rough decision of the classifier may encourage the negative transfer of target samples.
	
	Pan et al. \cite{pan2009survey} stated that a DA model is prone to negative transfers when it lags behind a non-DA model (which is trained only on the source domain) in performance. The fixed threshold ($0.5$) for constructing boundary between known and unknown target samples leads to biased adversarial learning. That is, for a target sample, when the classifier assigns an unknown probability low than the threshold based on its pseudo decision boundary, the generator will always be encouraged to align that sample towards known classes even if they belong to unknown classes. The OSBP method struggles to perform better than non-DA classifier for some tasks discussed in section \ref{results} because of negative transfers by aligning unknown target samples towards known classes. 
	
	To this end, we propose to extend their domain adversarial network by integrating a new multi-classifier based weighting module to the network. We refer to the extended generative model as the multi-classifier based adversarial domain adaptation model. To reduce negative transfers, first, we evaluate the underlying discriminative domain information of the known and unknown target samples, and then assign them with distinguishable weights which are more representative to their similarities to the source domain. In particular, the underlying domain information of target samples is measured by evaluating discriminative label information based on the resemblance to each source domain classes and probable similarity with the source domain classes when measured against the probability of belonging to the unknown classes of the target domain. This ensures that the proposed weighting module provides a rough estimate of the underlying domain of the target samples. Then based on our generated weights, the generative module performs adversarial domain adaptation and aligns known target samples towards known source samples and rejects the unknown target samples. Thus, in the proposed DA model, the generative module is not forced to draw a threshold driven boundary between known and unknown target samples, which may initiate negative transfers as the baseline method. The proposed model improves performance over the previous method \cite{saito2018open}, which indicates it constructs a good boundary between known and unknown target samples by eliminating negative transfers.
	Several attempts to address partial DA \cite{zhang2018importance, cao2019learning}, universal DA \cite{You2019UniversalDA}, and open set DA \cite{liu2019separate} by generating weights are identified in the literature. Zhang et al. \cite{zhang2018importance} utilize the output of an auxiliary domain discriminator to derive the probability of the source samples belonging to the target domain and assign weights to the source samples accordingly for partial DA. For improving prior partial DA methods participating in negative transfer \cite{cao2018partial, caopartial2018}, Example Transfer Network (ETN) \cite{cao2019learning} degrades the weight of source images from source private classes before integrating to the source classifier and places a discriminative domain classifier to quantify sample transferability. To generate weights for source samples belonging to the shared label sets, Universal Adaptation Network (UAN) \cite{You2019UniversalDA} integrates domain similarity and prediction uncertainty. Unlike our proposed method, UAN does not integrate underlying label information to evaluate domain similarity between the source and target samples. Above-mentioned methods concentrate on assigning weights to the source samples, whereas the OSBP DA setting demands to assign weights to the unlabeled  target samples. This is more challenging than weighting source samples as we do not know the labels of target samples and the shared label space during training. We propose a new multi-classifier based weighting scheme in an adversarial DA method for the OSBP DA setting. Separate to Adapt (STA) \cite{liu2019separate} method assigns weights to a target sample based on its highest similarity to one of the source domain classes. On the contrary, our proposed weighting module assesses domain information based on a target samples' similarity to known classes and dissimilarity to the unknown class, and then assign identifiable weights (please refer to Section \ref{weithing scheme} for details).\\
	The main contributions of this paper are as follows:
	
	\begin{itemize}
		\item We propose a domain adversarial model by integrating a new multi-classifier module in the OSBP domain adversarial model \cite{saito2018open} for the open set domain adaptation setting that has access to unknown classes only in the target domain. The multi-classifier structure introduces a weighting scheme in the proposed model, which assesses fundamental domain information based on distinctive label information for assigning identifiable weights to the known and unknown target samples. This enhances the positive transfer of target samples and facilitate the adversarial training. Unlike the previous method \cite{saito2018open}, which requires an assumption of the known-unknown boundary threshold, the proposed method is capable of automatically discovering the boundary between known and unknown target samples.
		\item We conduct comprehensive experiments and demonstrate that our proposed model reduces the rate of negative transfer and achieves better performance than contemporary domain adaptation methods on several datasets.
	\end{itemize}

	The proposed method can be applied to many multimedia applications such as image or video classification. Section II presents a brief discussion about contemporary DA methods. Details on our proposed method are described in Section III. Thorough performance comparison and analysis of the proposed method with contemporary DA approaches on different datasets are provided in Section IV.
	
	\section{Related Work}
	In this section, we briefly review recent domain adaptation methods based on the inter-domain relationship of label set constraints. These methods include those from closed set domain adaptation, partial domain adaptation, open set domain adaptation, and universal domain adaptation.
	\subsection{Closed Set Domain Adaptation}
	\label{cda}
	Closed set DA methods concentrate on reducing the divergence between the source and the target domains. Recent works have shown that because of its setting, closed set DA methods can exploit domain invariant features by explicitly reducing the domain divergence upon the supervised deep neural network structures. The development of deep learning based closed set DA methods \cite{long2018conditional, saito2018maximum, long2016unsupervised, tzeng2014deep, tzeng2017adversarial, long2015learning, ganin2016domain, haeusser2017associative, hoffman2017simultaneous} is originated from prior shallow DA methods \cite{saenko2010adapting, wang2014flexible, pan2010domain, gong2012geodesic, zhang2013domain, duan2012domain}. 
	Closed set DA methods fall into three main categories. The first category of methods is based on static moment matching, such as Maximum Mean Discrepancy (MMD) \cite{long2017deep, tzeng2014deep, long2016unsupervised, long2015learning}, Central Moment Discrepancy (CMD) \cite{zellinger2017central}, and second-order statistics matching \cite{sun2016deep}. The second category of methods adapts the adversarial loss concept of GAN \cite{goodfellow2014generative} and initiates the generation of images that are non-discriminative to the shared label space of source and target domain \cite{hoffman2017simultaneous, ganin2015unsupervised, tzeng2017adversarial}. Furthermore, domain adversarial methods align pixels and features from both domains and synthesize labeled target images for data augmentation \cite{volpi2018adversarial, sankaranarayanan2018generate, murez2018image, liu2018detach, huang2018auggan, bousmalis2017unsupervised}. Saito et al. \cite{saito2018maximum} utilize the probabilistic outputs of two classifiers’ to measure domain discrepancy loss and update both the classifiers based on this loss adversarially. However, we use two different types of classifiers and a discriminator for measuring underlying domain similarity and utilize this similarity to compute final domain discrepancy loss by updating only the main classifier. Recent research has explored Cycle-Consistent GAN \cite{zhu2017unpaired} for developing CycleGAN-based \cite{russo2018source, hoffman2017cycada, li2019cycle} DA methods. The final category of methods leverages Batch Normalization statistics for adapting domains to a canonical cone \cite{cariucci2017autodial, li2016revisiting}. 
	
	\subsection{Partial Domain Adaptation}
	The vanilla closed set DA's assumption that the source and target domains share the same label space does not hold in partial DA. The setting of partial DA assumes a target domain that has a fewer number of classes than the source domain. Cao et al. \cite{cao2018partial} use multiple domain discriminators along with class-level and instance-level weighting mechanism to obtain class-wise adversarial distribution matching for solving partial DA. Cao et al. \cite{caopartial2018} refine Selective Adversarial Network (SAN) \cite{cao2018partial} by using only one adversarial network and integrating the class-level weight to the source classifier. 
	\begin{figure*}[!ht]
		\centering
		\begin{subfigure}[b]{0.56\textwidth}
			\centering
			\includegraphics[width=.78\textwidth]{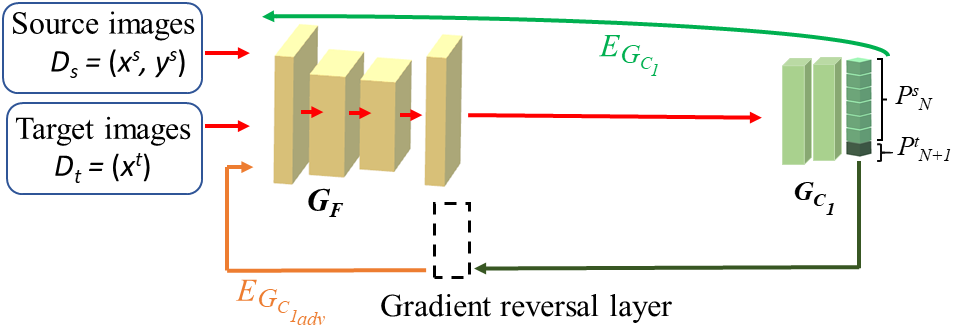}
			\caption{Block diagram of the OSBP domain adaptation model \cite{saito2018open}.}
			\label{f3}
		\end{subfigure}
		\begin{subfigure}[b]{0.7\textwidth}
			\centering
			\includegraphics[width=.8\textwidth]{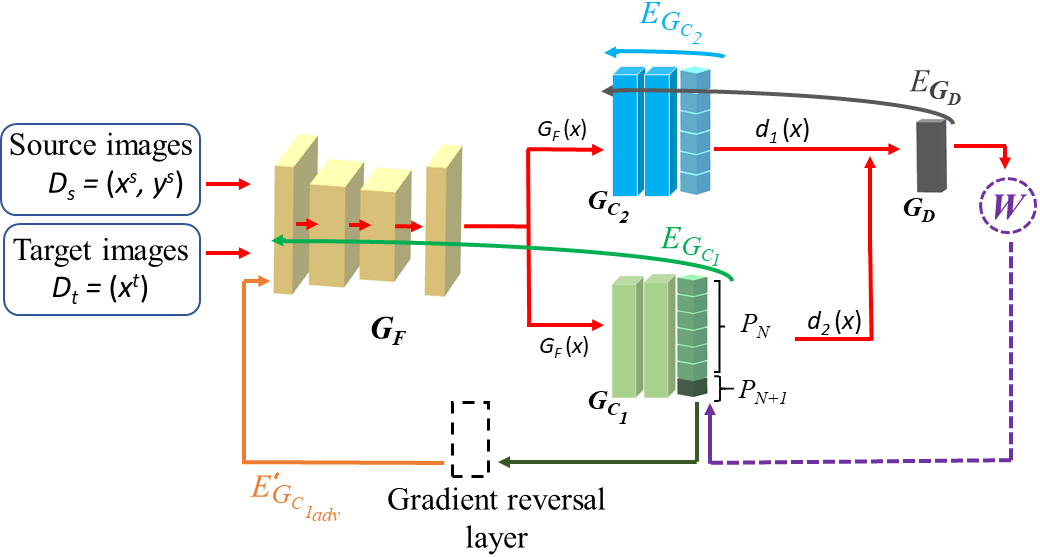}
			\caption{Training phase of the proposed domain adaptation model.}
			\label{f1} 
		\end{subfigure}
		\begin{subfigure}[b]{0.56\textwidth}
			\centering
			\includegraphics[width=.78\textwidth]{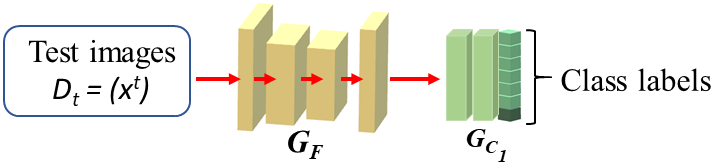}
			\caption{Testing phase of the proposed domain adaptation model.}
			\label{f2}
		\end{subfigure}
		
		\caption{(a) Block diagram of the baseline method \cite{saito2018open}. (b) Block diagram of the training phase of our proposed adversarial domain adaptation network with a multi-classifier based weighting scheme. Here, $G_F$, $G_{C_1}$, $G_{C_2}$ and $G_D$ denotes the generator, domain classifier, non-adversarial supplementary source and domain classifier respectively. $E_{G_{C_1}}$ and $E_{{G_{C_1}}_{adv}}$, $E_{G_{C_2}}$, and $E_{D}$ are errors for optimizing $G_F$ and $G_{C_1}$, $G_{C_2}$, and $G_D$ respectively. $G_{F}(x)$ stands for the generated features by the generator, which is fed into the domain classifier and supplementary source classifier simultaneously. $W$ is the generated weights for target samples. (c) A pictorial illustration of the proposed model during the testing phase. Red arrows denote forward passes while other arrows represent a backward pass.}
	\end{figure*}
	\subsection{Open Set Domain Adaptation}
	Open set DA methods aim to reject outliers or images from unknown classes while correctly recognizing inliers or images from the classes of interest (shared label space). Multi-class open set SVM \cite{jain2014multi} is designed to reject images from unknown classes. In this method, the SVMs are trained to assign probabilistic decision scores to samples and reject unknown samples by a threshold. Bendale et al. \cite{bendale2016towards} integrates an OpenMax layer upon deep neural networks for exploiting them in open set recognition. The OpenMax layer assists the network in estimating the probability of an image coming from an unknown class. To generate unknown samples for open set recognition, Ge et al. \cite{ge2017generative} combines a generative model with the OpenMax layer, and to reject unknown samples during testing, this method defines a threshold.
	
	As shown in Fig. \ref{1c}, Busto et al. \cite{panareda2017open} considers images from outliers (out of shared label space) of both source and target domains as `unknown' class. This method aligns target samples to source samples by an Assign-and-Transform-Iteratively (ATI) and trains SVMs for classification. This open set DA method takes advantage of the target samples from unknown classes to align with source samples from unknown classes. Saito et al. \cite{saito2018open} modified the open set DA setting for addressing more practical DA cases. Their modified setting does not require unknown classes in the source domain. They address this DA by adversarially training a classifier with an extra category named `unknown'. Unlike our method, OSBP does not explore any domain information instead enforces hard threshold and STA \cite{liu2019separate} depends only on the decision of source-only trained classifier for separating unknown target samples, which can give uncertain predictions for target samples. Our proposed method overcomes such shortcomings and improves classification performance over their methods.  

	\subsection{Universal Domain Adaptation}
	Recent research introduces a universal DA setting that imposes no previous knowledge on the source, and target domain label sets \cite{You2019UniversalDA}. This is another valuable step towards addressing a practical adaptation scenario. The authors proposed a Universal Adaptation Network (UAN), which combines domain similarity and prediction uncertainty while generating sample weights for finding shared label sets.
	
	\section{Proposed Methodology}
	In this section, we formally describe the open set domain adaptation setting, which is the focus of this work, discuss the limitations of the baseline method and present our proposed method.
	
	\subsection{Domain Adaptation Setting}
	The open set domain adaptation setting of our focus constitutes a source domain $D_s = {(x_i^s,y_i^s)}_{i=1}^{n_s}$ of $n_s$ labeled instances associated with $|C_s|$ classes, which are drawn from distribution $p_s$ and a target domain $D_t = {(x_j^t)}_{j=1}^{n_t}$ of $n_t$ unlabeled instances drawn from distribution $p_t$, where $p_s \neq p_t$. We denote the class labels of the target and source domains as $C_t$ and $C_s$ respectively. The shared label space is denoted as $C = C_s \cap C_t$. $\overline{C_t} = C_t \setminus C$ represents the label sets private to the target domain, which should be recognized as `unknown'. In this type of open set domain adaptation, it is difficult to identify which part of the target label space $C_t$ is shared with the source label space $C_s$ because the target domain is fully unlabeled and $C_t$ is unknown at the training time. It is challenging to differentiate between known and unknown target samples as we do not have any trace of the target sample labels.

	\subsection{Baseline Domain Adversarial Model}
	\label{baseline}
	In this section, we discuss in brief the OSBP \cite{saito2018open} model and its tendency to initiate negative transfers. 
	
	The OSBP method (Fig. \ref{f3}) is an adversarial DA model that aims to reduce divergence between the source and target domains by learning transferable features in a two-player minimax game in line with existing domain adversarial networks \cite{ganin2016domain, hoffman2017simultaneous}. The first player of the model is a domain classifier $G_{C_1}$ and the second player is a feature generator $G_F$. The final objective of the OSBP method is to correctly classify known target samples as corresponding known class and target samples belonging to unknown classes as `unknown'. 
	
	The feature generator $G_F$ takes inputs from both source domain $D_s$ and target domain $D_t$ at the same time. The domain classifier $G_{C_1}$ takes features from $G_F$ and outputs $N+1$ dimensional probability, where $N$ specifies the number of known or source categories ($C_s$) and the probability for the unknown category is indicated by the ${(N+1)}^{th}$ index. During the forward pass, within $G_{C_1}$, the features are transformed to a ${N+1}$-dimensional class probability through softmax function as, $\sigma({z}) = \exp{(z)}/(\sum_{i=1}^{N+1}\exp{(z_i)})$,
	where \textit{$z$} is the logit vector. 
	
	The OSBP method intends to construct a pseudo decision boundary for unknown classes. As the target domain is unlabeled during training, the domain classifier $G_{C_1}$ is weakly trained to construct a pseudo decision boundary between known source samples and target samples by putting the target samples on the side of the unknown category. The OSBP method trains the domain classifier $G_{C_1}$ to output $P(y = N+1|x_j^t) = T$ for `unknown class'. Then the feature generator $G_F$ is trained to deceive the domain classifier $G_{C_1}$ adversarially. The feature generator $G_F$ is trained with the ability to increase or decrease the `unknown' class probability $P(y = N+1|x_j^t)$ of the classifier $G_{C_1}$ for maximizing the error of $G_{C_1}$ and align target samples to known or unknown classes. Also, the OSBP method assumes that the empirical threshold value $T = 0.5$ dictates the generative model to construct a good boundary between known and unknown target samples.
	
	The OSBP method trains the domain classifier and the generator on source samples first as follows,
	\begin{equation} \label{e2}
	\begin{split}
	E_{G_{C_1}} & = \frac{1}{n_s}\sum_{i=1}^{n_s} L_{G_{C_1}}(G_F(x_i^s),y_i^s).\\
	\end{split}
	\end{equation}\\
	Here, $L_{G_{C_1}}$ is the standard cross-entropy loss function for minimizing the error of $G_{C_1}$. Then a binary cross-entropy loss is used for maximizing the error of $G_{C_1}$ adversarially to separate known and unknown target samples as follows,
	\begin{equation} \label{e3}
	\begin{split}
	E_{{G_{C_1}}_{adv}} =-\frac{1}{n_t}\sum_{j=1}^{n_t}T\log(P(y=N+1|x_j^t))\\
	-\frac{1}{n_t}\sum_{j=1}^{n_t} (1-T)\log(1-P(y=N+1|x_j^t)).\\
	\end{split}
	\end{equation}
	The overall training objective of the OSBP method is,
	\begin{equation} \label{e4}
	\begin{split}
	\theta_{G_{C_1}}&= \operatorname*{argmin}_{\theta_{G_{C_1}}} E_{G_{C_1}} + E_{{G_{C_1}}_{adv}}\\
	\theta_{G_F}&= \operatorname*{argmin}_{\theta_{G_F}} E_{G_{C_1}} - E_{{G_{C_1}}_{adv}}. 
	\end{split}
	\end{equation}
	The training objective indicates that the domain classifier $G_{C_1}$ tries to set `unknown' class probability $P(y = N+1|x_j^t)$ equal to $T$, on the other hand the generator $G_F$ tries to make $P(y = N+1|x_j^t)$ different from $T$ for maximizing the value of $E_{{G_{C_1}}_{adv}}$. For calculating the gradient of $E_{{G_{C_1}}_{adv}}$ efficiently, the OSBP method utilizes a gradient reversal layer proposed by \cite{ganin2015unsupervised}.\\
	Negative Transfer in the OSBP Method: The OSBP method does not evaluate any underlying domain level discerning characteristics but relies only on the pseudo decision of $G_{C_1}$. The lack of such domain knowledge and enforcement of an assumed boundary threshold $T = 0.5$ value for separating known and unknown classes will harm the model as the generator will attempt to align all the target samples with $P(y = N+1|x_j^t) < 0.5$ to known classes during training. As a result, the model will be deprived of the opportunity to learn such image features which should be recognized as unknown even when $P(y = N+1|x_j^t)$ is less than $0.5$. For example, during training, the domain classifier $G_{C_1}$ assigns $P(y = N+1|x_j^t) = 0.4$ for a target sample and the rest $0.6$ of the softmax probability is distributed over other $1, 2, ..., N$ indices of $G_{C_1}$ with no index holding $P \geq 0.4$. This probability outcome distribution suggests the sample should be aligned towards the unknown class. However, the generator will always find it easier to decrease $P(y = N+1|x_j^t) < 0.4$ to maximize the error of $G_{C_1}$ and align it towards known classes as it does not explore underlying domain knowledge before adversarial training. Thus, the model will be exposed to negative transfers during training and testing i.e., target samples from unknown classes will be aligned to known classes.
	
	To reduce the propensity for such negative transfers in the OSBP method and improve classification performance, we propose to extend their domain adversarial model. Our proposed multi-classifier based domain adversarial model is discussed in the next section.
	
	\subsection{Proposed Multi-classifier Based Domain Adversarial Model}
	The limitations of the OSBP method is mainly due to the lack of an indicator of the likelihood of a target sample belonging to known or unknown classes. This prompted us to design a method to produce the indicator of target samples belonging to known or unknown classes beforehand the adversarial training to facilitate DA. The proposed method is illustrated in Fig. \ref{f1}. To investigate underlying domain information of target samples for reducing negative transfers, we propose to integrate a multi-classifier based weighting module in the baseline network. Our proposed method comprises of two modules: 1)Adversarial module and 2) Multi-classifier based weighting module. 
	\subsubsection{Adversarial module}
	The adversarial module of our proposed method has a similar structure to the OSBP method. The first player of the model is a domain classifier $G_{C_1}$ which is trained to distinguish the features of the source domain from the target domain. The second player is a feature generator $G_F$ which is simultaneously trained to reduce feature distribution divergence in the opposite direction of the domain classifier. However, our proposed method follows a different adversarial optimization procedure for distinguishing unknown target samples from the known target samples compared to the OSBP. The ultimate goal is to train a source domain classifier that is transferred to the target domain classifier with an extra category named `unknown'.
	
	The proposed weighting module quantifies each target sample with a generated weight $W(x_j^t)$ (Section \ref{weithing scheme}), which is an encoding of the underlying discriminative domain information. In particular, prior to adversarial DA, we assign identifiable weights $W(x_j^t)$ to known and unknown target samples based on their similarity to source domain to facilitate the generator in deciding whether to decrease or increase the `unknown' class probability $P(y = N+1|x_j^t)$ for maximizing the error of $G_{C_1}$ and eventually align the target samples to known classes or `unknown' class. Therefore, the generator does not have to draw an empirical threshold driven boundary between known-unknown target samples by depending only on the pseudo decision of the classifier, which may encourage negative transfer. 
	
	We use the standard cross-entropy loss optimization, as illustrated in Equation (\ref{e2}) for minimizing the error of the domain classifier $G_{C_1}$. To distinguish known from unknown target samples and simultaneously maximize the error of the domain classifier $G_{C_1}$ adversarially, we use a binary cross-entropy loss. We infuse our computed weight measure for the target samples in the optimization procedure as follows,
	\begin{equation} \label{e5}
	\begin{split}
	{E^{'}_{{G_{C_1}}_{adv}}} =-\frac{1}{n_t}\sum_{j=1}^{n_t}W(x_j^t)\log(P(y=N+1|x_j^t))\\
	-\frac{1}{n_t}\sum_{j=1}^{n_t} (1-W(x_j^t))\log(1-P(y=N+1|x_j^t)).\\
	\end{split}
	\end{equation}
	The mini-max game between the generator and domain classifier in the adversarial model is equivalent to aligning target samples towards known classes of the source domain or `unknown' class based on their weights. 

	\subsubsection{Multi-classifier based weighting module}
	\label{weithing scheme}
	In this section, we present the conceptual details of the weighting scheme of our proposed DA method. 
	
	\textbf{Overview:} The main challenge in our proposed method, as illustrated in Equations (\ref{e2}) and (\ref{e5}), is the way of measuring the probability of each target samples resembling the source domain on the basis of which the boundary between known and unknown classes is to be constructed. We aim to develop a weight measure $W(x_j^t)$ for each target samples based on their discriminative label and domain information. We propose to integrate a supplementary source classifier $G_{C_2}$ in the adversarial model (Fig. \ref{f1}) to determine the similarity of target samples to individual known-source labels $C_s$. The combined similarity to each known classes represents the similarity to the source domain. And, utilize the pseudo-decision of $G_{C_1}$ to compute the combined similarity to all known classes (i.e., similarity to source domain) measured against target private labels $\overline{C_t}$. This ensures the exploitation of probable underlying label information of target samples concerning both known and unknown classes. 
	
	We further introduce a non-adversarial supplementary domain classifier $G_D$ in the model to evaluate underlying domain information and produce a weight for target samples. The non-adversarial supplementary domain classifier $G_D$ assumes that the target samples belonging to the shared label space $C$ are closer to the source domain samples than $\overline{C_t}$. Being inspired by a prior work \cite{odena2017conditional} that integrated the label information into the domain discriminator, we combine the domain similarity measure of $G_{C_1}$ and $G_{C_2}$, and encode in $G_D$. Now that the supplementary domain classifier $G_D$ holds the encoded information required to serve our purpose, $G_D$ is jointly trained with $G_{C_2}$ to distinguish the source domain samples from the target domain samples by utilizing the Sigmoid probability of classifying each target sample ($x_j^t$) to the source domain. The output ($W(x_j^t)$) of $G_D$ for target samples gives the probability of a sample belonging to the shared label space.
	
	
	This constitutes our multi-classifier based weighting module. The purpose of the weighting scheme is to assign either high or low weights to the target samples depending on their similarity to the source domain and reduce the chances of negative transfer. Generated weights are back-propagated to the adversarial module for optimizing the generator $G_F$ and the domain classifier $G_{C_1}$ to construct boundary between known and unknown classes. 
	
	
	\textbf{Mechanism in detail: } We place the supplementary source classifier $G_{C_2}$ to predict the source class labels with a leaky-softmax function \cite{cao2019learning}, which maintains the total probability of less than 1. The supplementary source classifier $G_{C_2}$ converts features of the generator $G_F$ to $|C_s|$-dimensional class probabilities as follows,
	\begin{equation} \label{e6}
	\begin{split}
	\overline{\sigma}(l) & = \frac{\exp{(l)}}{|C_s| + \sum_{c=1}^{|C_s|}\exp{(l_c)}}\\
	\end{split}
	\end{equation}
	where $l$ is the logit vector. The parameters of $G_{C_2}$ is trained only on the source samples; therefore, unlike source samples, target samples will have smaller logits or uncertain predictions. We define the probability of each sample belonging to the source domain based on known-source label information $d_1(x)$ as follows,
	\begin{equation} \label{d1}
	\begin{split}
	d_1(x) = \sum_{k=1}^{|C_s|}G_{C_2}^k(G_F(x))\\
	\end{split}
	\end{equation}
	where, $G_{C_2}^k(G_F(x))$ is the probability of a sample belonging to the $k^{th}$ known class. The element-sum ($d_1(x)$) of the leaky-softmax outputs for samples resembling the source domain will be high or close to 1 whereas, samples dissimilar to source domain will yield low or close to 0 outputs. 
	That is, the higher the value of $d_1(x_j^t)$ is, the higher the chance that a target sample lies in the vicinity of shared label space $C$. On the other hand, the smaller the value of $d_1(x_j^t)$ is, the more probable that the target sample comes from $\overline{C_t}$. We train $G_{C_2}$ by a multiclass one-vs-rest binary loss for the $|C_s|$-class classification as, 
	\begin{equation} \label{e7}
	\begin{split}
	E_{G_{C_2}} = -\frac{1}{n_s} \sum_{i=1}^{n_s}\sum_{k=1}^{|C_s|}y_{i,k}^s\log G_{C_2}^k (G_F(x_i^s))\\
	+(1 - y_{i,k}^s) \log(1 - G_{C_2}^k(G_F(x_i^s)))\\
	\end{split}
	\end{equation}\\
	where $y_{i,k}^s$ denotes the ground-truth label for source example $x_i^s$ and the probability of each sample $x$ belonging to class $k$ is $G_{C_2}^k(G_F(x))$. This similarity measure defined so far is still exposed to risk as to the value of $d_1(x)$ for target samples can be uncertain. To further support the similarity measure, we compute the probability of a sample belonging to $C$ when measured against the `unknown' class probability from the domain classifier $G_{C_1}$ as follows,
	\begin{equation} \label{d2}
	\begin{split}
	d_2(x) = (1 - P(y = N+1|x)).\\
	\end{split}
	\end{equation}
	Target samples residing in the shared label set are likely to produce higher $d_2(x)$ than unknown target samples. Now, we define our final similarity measure supported by $G_{C_1}$ and $G_{C_2}$ as,
	\begin{equation} \label{e8}
	\begin{split}
	G_D(G_F(x))& = (d_2(x))(d_1(x)).\\
	\end{split}
	\end{equation}
	Now, $G_D(G_F(x_j^t))$ can be seen as the complete measure of the likelihood of target samples belonging to shared label space $C$ i.e., for target samples, the higher the value of $G_D(G_F(x))$ is the more probable that it belongs to the shared classes. \\
	For the convenience of understanding, we represent the possible cases of outcomes from the two deciding factors (i.e., $d_1(x_j^t)$ and $(d_2(x_j^t)$) for computing the similarity measure of the known and unknown target samples as follows:\\
	\textbf{$A\textsubscript{H}$}: For the target samples belonging to the shared label space $C$, it is highly likely the output of $d_1(x_j^t)$ will be high.\\
	\textbf{$A\textsubscript{L}$}: For the target samples belonging to the target private label space $\overline{C_t}$, it is highly likely the output of $d_1(x_j^t)$ will be low.\\
	\textbf{$B\textsubscript{H}$}: For the target samples belonging to the shared label space $C$, it is highly likely the output of $d_2(x_j^t)$ will be high.\\
	\textbf{$B\textsubscript{L}$}: For the target samples belonging to the target private label space $\overline{C_t}$, it is highly likely the output of $d_2(x_j^t)$ will be low.
	Note that, here, high means close to 1 and low means close to 0. 
	In the cases below, ($A, B$) denotes the occurrence of $A$ and $B$ for measuring the similarity of a target sample to the source domain.\\
	\textbf{Case 1 ($A_L, B_L$):} For this case, the computed similarity measure (Equation (\ref{e8})) will be low and this will assist the generator in deciding to increase the value of $P(y = N+1|x_j^t)$ for maximizing the error of the domain classifier $G_{C_1}$ and align the sample towards the `unknown' class.\\
	\textbf{Case 2 ($A_L, B_H$):} For this case, the computed similarity measure (Equation (\ref{e8})) will be low and this will assist the generator in deciding to increase the value of $P(y = N+1|x_j^t)$ for maximizing the error of the domain classifier $G_{C_1}$ and align the sample towards the `unknown' class.\\
	\textbf{Case 3 ($A_H, B_L$):} For this case, the computed similarity measure (Equation (\ref{e8})) will be low and this will assist the generator in deciding to increase the value of $P(y = N+1|x_j^t)$ for maximizing the error of the domain classifier $G_{C_1}$ and align the sample towards the `unknown' class.\\
	\textbf{Case 4 ($A_H, B_H$):} For this case, the computed similarity measure (Equation (\ref{e8})) will be high and this will assist the generator in deciding to decrease the value of $P(y = N+1|x_j^t)$ for maximizing the error of the domain classifier $G_{C_1}$ and align the sample towards the known classes. 

	We train the supplementary domain classifier $G_D$ as follows,
	\begin{equation} \label{e9}
	\begin{split}
	E_{G_D} = -\frac{1}{n_s} \sum_{i=1}^{n_s}\log (G_D(G_F(x_i^s))\\ -\frac{1}{n_t} \sum_{j=1}^{n_t} \log (1-G_D(G_F(x_j^t))).\\  
	\end{split}
	\end{equation}
	Equations (\ref{e8}) and (\ref{e9}) indicate that the outputs of $G_D$ are dependent on the output of the supplementary source classifier $G_{C_2}$ and the output of the domain classifier $G_{C_1}$. This verifies that $G_D$ is trained to evaluate target samples based on the discriminative known classes and unknown class label information, which will assist $G_D$ to assign meaningful and identifiable weights to target samples belonging to $C$ and $\overline{C_t}$. Thus, we obtain weights to quantify the similarity of target samples to the source domain from $G_D$ as,
	\begin{equation} \label{e10}
	\begin{split}
	W(x_j^t) & = G_D(G_F(x_j^t).\\
	\end{split}
	\end{equation}
	During the early phase of training, if either one of the classifiers ($G_{C_1}$, $G_{C_2}$) produces uncertain similarity measure ($d_2(x)$, $d_1(x)$), it will be supported by the other ones’ decision for assigning weights to the target samples. However, over the training epochs, both the classifiers will converge to their optimal value for the feature extractor $G_F$. In such an advanced phase of training, the target samples belonging to the shared label space will surely get close to 1 similarity score from $G_{C_2}$ as it is trained on source samples only. Similarly, the similarity score for that target sample from $G_{C_1}$ will also be high as $G_{C_1}$ learns to yield low `unknown’ class probability for target samples belonging to $C$. Thus, the combined weight $W(x_j^t)$ will be high, and the sample will be aligned to the known classes. On the other hand, if a target sample comes from outside the shared label space, then the $G_{C_2}$ will produce close to 0 similarity score. The $G_{C_1}$ will produce low score as well for such sample, and eventually, the weight $W(x_j^t)$ will be low, and the sample will be aligned to the `unknown’ class. Section \ref{weights} provides pictorial representation of learned weights.
	
	Considering all the above-discussed derivations, we present our proposed adversarial domain adaptation model with multi-classifiers. We denote the parameters of the supplementary source classifier $G_{C_2}$ as $\theta_{G_{C_2}}$. The overall objectives of our proposed method are:
	\begin{equation} \label{e11}
	\begin{split}
	\theta_{G_{C_1}} & = \operatorname*{argmin}_{\theta_{G_{C_1}}} E_{G_{C_1}} + E^{'}_{{G_{C_1}}_{adv}}\\
	\theta_{G_F} & = \operatorname*{argmin}_{\theta_{G_F}} E_{G_{C_1}} - E^{'}_{{G_{C_1}}_{adv}}\\
	\theta_{G_{C_2}} & = \operatorname*{argmin}_{\theta_{G_{C_2}}} E_{G_{C_2}} + E_{D}.\\
	\end{split}
	\end{equation}
	During back-propagation, we use a gradient reversal layer \cite{ganin2015unsupervised} to calculate the gradient of $E^{'}_{{G_{C_1}}_{adv}}$ efficiently.
	Unlike prior work \cite{saito2018open}, our proposed model does not need any prior training on the source dataset. We optimize all the objectives simultaneously in an end-to-end fashion.
	
	\subsubsection{Constraining positive transfer}
	In this section, we discuss how the proposed method limits the tendency of negative transfer in the OSBP model based on the case discussed in Section \ref{baseline}, which explains negative transfers in the OSBP model. In contrary to OSBP model, for example, when the domain classifier assigns $P(y = N+1|x_j^t) =0.4$ for a target sample, the $G_D$ in our proposed method will generate a weight $W(x_j^t)$ for the sample after evaluating its similarity to the source domain based on Equations (\ref{e6} - \ref{e10}). In short, we compute the value of $d_2(x_j^t)$ (Equation (\ref{d1})) and $d_1(x_j^t)$ (Equation (\ref{d1})). The former one is $0.6$ in this case (we consider this value as a high as it is closer to 1 than 0), and the latter one can be either high and yield high weight $W(x_j^t)$ or low leading to low weight $W(x_j^t)$ based on Equation (\ref{e10}). If the weight $W(x_j^t)$ is high, the proposed model will assist the generator in aligning the sample to known classes by decreasing $P(y = N+1|x_j^t)$. Otherwise, if the weight $W(x_j^t)$ is low, the sample will be aligned to `unknown' class by maximizing the value of $P(y = N+1|x_j^t)$. Thus, our DA model does not participate in negative transfer by aligning unknown samples to known classes. 
	
	\subsubsection{Testing Phase}
	During the training phase, we fulfill our goal to transform the domain classifier $G_{C_1}$ from source domain classifier to target domain classifier, including the category `unknown' by utilizing $G_{C_1}$ and $G_D$ classifiers. In the testing phase, we omit the supplementary classifiers and utilize only the trained feature generator $G_F$, and $G_{C_1}$ to classify test images correctly, as shown in Fig. \ref{f2}.
	
	\begin{table*}[!ht]
		\caption{Classification accuracy (\%) of proposed and contemporary domain adaptation methods on Office-Home dataset tasks.}
		\vspace{-4mm}
		\label{t1}
		\begin{center}
			\resizebox{\textwidth}{!}{%
				\begin{tabular}{ccccccccccccccccccccccccccc}
					\hline
					\multirow{3}{*}{Approach} & \multicolumn{26}{c}{Accuracy (\%)}\\\cline{2-27}
					&\multicolumn{2}{c}{Ar$\rightarrow$ Cl} & \multicolumn{2}{c}{Ar$\rightarrow$ Pr} & \multicolumn{2}{c}{Ar$\rightarrow$ Rw} & \multicolumn{2}{c}{Cl$\rightarrow$ Ar} & \multicolumn{2}{c}{Cl$\rightarrow$ Pr} & \multicolumn{2}{c}{Cl$\rightarrow$ Rw} & \multicolumn{2}{c}{Pr$\rightarrow$ Ar} & \multicolumn{2}{c}{Pr$\rightarrow$ Cl} & \multicolumn{2}{c}{Pr$\rightarrow$ Rw} & \multicolumn{2}{c}{Rw$\rightarrow$ Ar} & \multicolumn{2}{c}{Rw$\rightarrow$ Cl} & \multicolumn{2}{c}{Rw$\rightarrow$ Pr} &\multicolumn{2}{c}{Avg}\\ 
					& OS& OS$^\star$& OS& OS$^\star$& OS& OS$^\star$& OS& OS$^\star$& OS& OS$^\star$& OS& OS$^\star$& OS& OS$^\star$& OS& OS$^\star$& OS& OS$^\star$& OS& OS$^\star$& OS& OS$^\star$& OS& OS$^\star$& OS& OS$^\star$\\\hline\hline
					\rowcolor{lightgray}ResNet \cite{he2016deep} (2016)&54.4&54.8&69.5&70.1&78.7&78.1&62.0&61.3&60.8&62.3&71.6&72.5&64.2&64.4&58.9&58.6&75.5&76.1&70.3&69.2& 52.5&51.5&74.5&75.3&66.1&66.2\\
					DANN \cite{ganin2016domain} (2016) &-&\textit{44.8}&-&\textit{68.5} & -& 79.5& -&65.5&-&\textit{57.9}&-&\textit{67.4}&-& \textit{56.9}&-&\textit{40.2}&-& 77.5&-& \textit{68.5}&-&\textit{45.2}&-& 77.6&-&\textit{62.4}\\
					RTN \cite{long2016unsupervised} (2016)& -&\textit{50.9}&-& 75.6 &- & 82.9&- & 66.5&-&73.4 &-& 85.7&- & 65.6 &-&\textit{47.9} & -& 84.5& -& 78.1& -& 56.9& -& 77.6&-&70.4\\
					IWAN \cite{zhang2018importance} (2018)&\textit{53.1}&\textit{52.1}& 79.4&78.5& 86.1&86.4 &70.2&69.7 &70.9&71.3& 86.8&85.1& 74.9&74.5&\textit{55.6}&\textit{55.7}& 85.1&84.2& 77.9&78.7& 60.8&59.4& 77.2&76.8&73.1&72.7\\
					ETN \cite{cao2019learning} (2019)& 59.3&59.0& 77.1&76.8& 85.6&85.7& 63.1&62.9& 65.6&65.1& 75.3&75.6& 68.3&67.8&\textit{55.4}&\textit{55.6}& 86.4&85.9& 78.7&77.5& 62.3&61.5& 84.4& 84.2&71.8&71.4\\
					UAN \cite{You2019UniversalDA} (2019) &63.0&62.5& 82.8&82.4& 86.8&85.9& 76.8&76.9& 78.7&79.1& 84.4&84.8& 78.2&77.4&\textit{58.6}&\textit{57.8}& 86.8&85.9& 83.4&82.5& 63.2&63.0& 79.1&78.1&76.8&76.6\\
					BP \cite{ganin2015unsupervised} (2015) &\textit{53.6}&\textit{51.0} &\textit{69.1}&\textit{65.9} &\textit{75.9}&\textit{74.1} &59.5&57.3 &65.2&62.5 &73.2&\textit{72.0} &\textit{47.2}&\textit{45.0} &\textit{43.9}&\textit{40.2}&78.7&76.4 &70.6&\textit{65.3}&\textit{45.6}&\textit{42.1}&77.5&\textit{74.2}& \textit{63.3}&\textit{60.5}\\ 
					ATI \cite{panareda2017open} (2017)&\textit{53.8}&\textit{51.3}& 80.4&77.9& 86.1&85.0& 71.2&67.8& 72.3 &70.5&85.1&83.2 &74.3&72.5 &\textit{57.9}&\textit{55.1} &85.6 &84.7&76.1&75.2 &60.2&58.7 &78.3&77.0&73.4&71.5 \\ 
					OSBP \cite{saito2018open} (2018)&\textit{48.5}&\textit{48.6}& 70.9&70.6& \textit{75.2}&\textit{74.2}&59.5&58.2&61.6&\textit{59.9}&75.1&74.5&\textit{61.9}&\textit{62.2}&\textit{43.5}&\textit{43.2}&79.9&80.4&\textit{70.1}&70.2&53.9&54.1&75.7&75.4&\textit{64.6}&\textit{64.3}\\ 
					STA \cite{liu2019separate} (2019)&57.8&58.1& 71.3&70.1& 84.9&85.5& \textit{61.4}&61.9& 68.1&67.9& 75.2&75.8& 64.3&\textit{63.2}& \textit{51.8}&\textit{52.2}& 80.2&79.1& 73.9&74.2& 53.6&54.5& 80.5&81.4&68.6&68.7\\
					Our(ResNet-50) &64.8&64.9 &84.6& 84.9&\textbf{88.1}&\textbf{88.2} &79.6&79.9 &81.6&82.9 & 85.6&86.8& 77.9&78.5&61.8&63.1 &85.9& 87.5&85.4&85.6 & 67.5&65.2& 80.9&80.1&78.6&79.0\\			
					Our(ResNet-152) &\textbf{66.1}&\textbf{66.9} &\textbf{85.9} &\textbf{86.7}&87.6&87.9 &\textbf{80.5}&\textbf{80.8} &\textbf{83.5}&\textbf{85.6} &\textbf{86.9} &\textbf{87.9}&\textbf{79.9}&\textbf{81.1} &\textbf{63.1}&\textbf{64.8}&\textbf{87.9}&\textbf{88.7} &\textbf{86.9}&\textbf{88.2}&\textbf{68.8}&\textbf{69.7} &\textbf{82.1}&\textbf{82.8}&\textbf{80.0} &\textbf{80.9}\\\hline		\end{tabular}
			}
		\end{center}
	\end{table*}
	\begin{table*}[!h]
		\caption{Classification accuracy (\%) of proposed and other domain adaptation methods on Office-31 tasks.}
		\vspace{-4.5mm}
		\label{t2}
		\begin{center}
			\resizebox{\textwidth}{!}{%
				\begin{tabular}{ccccccccccccccc}
					\hline
					\multirow{3}{*}{Approach} & \multicolumn{14}{c}{Accuracy (\%)}\\\cline{2-15}
					& \multicolumn{2}{c}{A$\rightarrow$ W} & \multicolumn{2}{c}{D$\rightarrow$ W} & \multicolumn{2}{c}{W$\rightarrow$ D} & \multicolumn{2}{c}{A$\rightarrow$ D} & \multicolumn{2}{c}{D$\rightarrow$ A} & \multicolumn{2}{c}{W$\rightarrow$ A} & \multicolumn{2}{c}{Avg}\\ 
					& OS& OS$^\star$& OS& OS$^\star$& OS& OS$^\star$& OS& OS$^\star$& OS& OS$^\star$& OS& OS$^\star$& OS& OS$^\star$\\\hline\hline
					\rowcolor{lightgray}ResNet \cite{he2016deep} (2016)&83.2&83.8&93.8&94.7&95.8&94.6&84.6&84.7&72.3&71.9&75.5&75.3&84.2&84.2\\
					DANN \cite{ganin2016domain} (2016)&-&\textit{80.2}&-&\textit{79.9}&-&\textit{87.5}&-&\textit{80.4}&-&74.9&-&80.9&-&\textit{80.6}\\
					RTN \cite{long2016unsupervised} (2016)&-& 88.1& -&\textit{89.8}& -&\textit{84.8} &-& \textit{73.1} &-&85.1 & -&84.7 &-&84.2\\
					IWAN \cite{zhang2018importance} (2018) &86.5&84.9& \textit{89.9}&\textit{87.1}& \textit{91.1}&\textit{90.3}& \textit{82.3}&\textit{79.6}& 82.2&80.6& 85.7&83.1& 86.3&84.3\\
					ETN \cite{cao2019learning} (2019)& 85.7&84.4 & \textit{93.6}&\textit{92.9}& 96.9&96.0& 84.9&85.1& 84.9&85.6& 85.2&85.0&88.5&88.1\\
					UAN \cite{You2019UniversalDA} (2019) & 85.6&84.3& 94.7&\textit{93.9}& 97.9&97.5& 86.5&84.9& 85.5&85.6& 85.1&84.6& 89.2&88.4\\
					BP \cite{ganin2015unsupervised} (2015)&\textit{75.9}&\textit{74.1}&\textit{89.7}&\textit{87.2}&\textit{94.4}&\textit{93.2} &\textit{78.4}&\textit{76.8}&\textit{56.8}&\textit{55.3}&\textit{62.9}&\textit{62.7}&\textit{76.3}&\textit{74.8} \\ 
					ATI \cite{panareda2017open} (2017) &\textit{78.4}&\textit{74.9}& \textit{92.6}&\textit{90.6} &97.1&95.6 &\textit{78.9}&\textit{77.5}& \textit{71.6}&\textit{70.1} &76.8&\textit{74.2} &\textit{82.5}&\textit{80.4}\\ 
					OSBP \cite{saito2018open} (2018)& \textit{67.4}&\textit{66.9}& \textit{83.7}&\textit{83.5}& \textit{95.1}&94.8& \textit{82.6}&\textit{82.0}& 76.6&76.5& 79.5&78.9&\textit{80.8}&\textit{80.4}\\
					STA \cite{liu2019separate} (2019)& 88.6&90.1& 97.1&95.2& 97.3&97.5& \textbf{91.9}&\textbf{93.1}& 88.3&88.6& 84.1&84.2& 91.2&91.4\\
					Our(ResNet-50) &88.3&88.4 &97.3 &97.8&98.1 &98.4&87.8 &88.6&89.9 &89.6&84.9&85.8 & 91.1&91.5\\
					Our(ResNet-152) &\textbf{90.2}&\textbf{90.8}&\textbf{97.9} &\textbf{98.8}&\textbf{98.6}&\textbf{98.8} &89.5&89.7 &\textbf{91.0}&\textbf{91.7} &\textbf{86.8}&\textbf{87.6} & \textbf{92.4}&\textbf{92.9}\\\hline
				\end{tabular}
			}
		\end{center}
	\end{table*}
	\begin{table}[!h]
		\caption{Classification accuracy (\%) of proposed and other domain adaptation methods on ImageNet-Caltech tasks.}
		\vspace{-4.5mm}
		\label{t3}
		\begin{center}
			\resizebox{\columnwidth}{!}{%
				\begin{tabular}{ccccccc}
					\hline
					\multirow{3}{*}{Approach} & \multicolumn{6}{c}{Accuracy (\%)}\\\cline{2-7}
					& \multicolumn{2}{c}{I$\rightarrow$ C} & \multicolumn{2}{c}{C$\rightarrow$ I}&\multicolumn{2}{c}{Avg}\\ 
					& OS& OS$^\star$& OS& OS$^\star$& OS& OS$^\star$\\\hline\hline
					\rowcolor{lightgray}ResNet \cite{he2016deep}&75.7&75.1&67.1&67.8&71.4&71.5\\ 
					DANN \cite{ganin2016domain} (2016) & -& \textit{71.1}&-& \textit{65.9}&-&\textit{68.5}\\
					RTN \cite{long2016unsupervised} (2016)&- & \textit{71.9}&-& \textit{66.2} &-&\textit{69.1}\\
					IWAN \cite{zhang2018importance} (2018)& \textit{74.1}&\textit{72.9}& 68.7&\textit{65.3}&71.4&\textit{69.1}\\
					ETN \cite{cao2019learning} (2019)& 74.9&\textit{74.8}& 69.8&69.9&72.4&72.4\\
					UAN \cite{You2019UniversalDA} (2019)& \textit{75.3}&76.3& 70.2&70.8&72.7&73.6\\
					BP \cite{ganin2015unsupervised} (2015) &\textit{68.9}&\textit{67.3}&\textit{61.2}&\textit{59.0}&\textit{65.0}&\textit{63.2}\\
					ATI \cite{panareda2017open} (2017)&\textit{71.6}&\textit{65.9}& 67.4&\textit{65.1}&\textit{69.5}&\textit{65.5}\\ 
					OSBP \cite{saito2018open} (2018) &\textit{63.1}&\textit{63.4}&\textit{54.8}&\textit{53.6}&\textit{58.9}&\textit{58.5}\\ 
					STA \cite{liu2019separate} (2019) &75.3&\textit{74.2}&68.1&68.3&71.7&\textit{71.3}\\ 
					Our(ResNet-50)& 77.4&77.8& 69.8&70.1&73.6&74.0\\
					Our(ResNet-152)& \textbf{78.9}&\textbf{79.7}& \textbf{71.9}&\textbf{71.7}&\textbf{75.4}&\textbf{75.7}\\\hline
				\end{tabular}
			}
		\end{center}
	\end{table}
	\begin{table*}[!h]
		\caption{Classification accuracy (\%) of proposed and other domain adaptation methods on VisDA2017 tasks.}
		\vspace{-4.5mm}
		\label{t4}
		\begin{center}
			\resizebox{.88\textwidth}{!}{%
				\begin{tabular}{cccccccccc}
					\hline
					\multirow{3}{*}{Approach} & \multicolumn{9}{c}{Accuracy (\%)}\\\cline{2-10}
					& bicycle& bus & car &motorcycle &train &truck&unknown& OS& OS$^\star$\\\hline\hline
					\rowcolor{lightgray}ResNet \cite{he2016deep} (2016)
					&40.2 &55.4&63.5&70.8&74.1&35.2&45.6&54.9&56.5\\
					DANN \cite{ganin2016domain} (2016) &	\textit{32.4}&\textit{51.6}& 65.1& 71.3& 85.1&	\textit{23.1}&-&-&\textit{52.1}\\
					RTN \cite{long2016unsupervised} (2016)&	\textit{31.6}&63.6&\textit{54.2}&76.9&\textbf{87.3}&\textit{21.5}&-&-& \textit{51.1}\\
					IWAN \cite{zhang2018importance} (2018)&	\textit{30.6}&69.8 &\textit{58.3} &76.8 &	\textit{65.5}&\textit{30.8}&69.7&57.3&	\textit{55.3}\\
					ETN \cite{cao2019learning} (2019)&	\textit{31.6}&66.8 &\textit{61.7}&77.8 &	\textit{70.8}&\textit{30.8} &70.7&58.6&56.6\\
					UAN \cite{You2019UniversalDA} (2019)& 42.6&67.8&65.7&76.9&\textit{69.8}&	\textit{31.8}&70.7&60.9&59.1\\
					BP \cite{ganin2015unsupervised} (2015)&\textit{31.8}&66.5 &\textit{50.5}& 70.1&86.9 &\textit{21.8}&\textit{38.5}&	\textit{52.3}&\textit{54.6}\\ 
					ATI \cite{panareda2017open} (2017) &\textit{33.6}&\textit{51.6}&64.2 &78.1 &85.3 &\textit{22.5}&	\textit{42.5}&	\textit{54.8}&\textit{52.6}\\ 
					OSBP \cite{saito2018open} (2018)
					&\textit{35.6}&59.8 &\textit{48.3}&76.8 &	\textit{55.5}&\textit{29.8}&81.7 &55.4&	\textit{50.9}\\ 
					STA \cite{liu2019separate} (2019)
					&50.1 &69.1 &\textit{59.7}&\textbf{85.7} &84.7&\textit{25.1}&\textbf{82.4} &65.3&62.4\\
					Our(ResNet-50)&50.6&74.8&66.7&80.6&75.9&38.8&73.9&65.9&64.6\\
					Our(ResNet-152)& \textbf{52.1}&\textbf{77.7}&\textbf{67.7}&81.4&80.8&\textbf{39.8}&75.5&\textbf{67.8}&\textbf{66.6}\\\hline
				\end{tabular}
			}
		\end{center}
	\end{table*}
	
	\section{Experimental Studies}
	In this section, we describe the datasets, our evaluation details, and the results. We conduct experiments to evaluate our proposed method with contemporary DA methods on four standard datasets.
	\subsection{Datasets} 
	\textbf{Office-31} \cite{saenko2010adapting} has 31 categories in three visually distinct domains, namely: amazon (\textbf{A}), DSLR (\textbf{D}) and webcam (\textbf{W}). This dataset comprises a collection of samples from \textbf{amazon.com}, captured samples from DSLR and web camera for DA. We have chosen the first 10 classes as $C$ and the last 10 classes as `unknown' samples in the target domain $\overline{C_t}$ for accomplishing six open set DA tasks: A$\rightarrow$ W,  D$\rightarrow$ W, W$\rightarrow$ D, A$\rightarrow$ D, D$\rightarrow$ A and W$\rightarrow$ A. 
	\textbf{VisDA2017} \cite{peng2018visda} poses a special DA setting by focusing on a simulation (rendered 3D images) to real-world DA setting. Game engines generate the samples of source domain while the target domain samples are actual images. This dataset comprises 12 categories. Inline with \cite{saito2018open}, we have chosen six classes (bicycle, bus, car, motorcycle, train, and truck) as the shared classes $C$ and the remaining six classes as `unknown' classes in the target domain. 
	
	\textbf{Office-Home} \cite{venkateswara2017deep} consists of 65 classes in four different domains: Artistic images (Ar), Clip-Art images (Cl), Product images (Pr), and Real-World images (Rw). The first 10 classes in alphabetical order are used as the shared classes $C$. Leaving the next five classes private to the source domain, the rest classes are considered as `unknown' or private to the target domain. For this dataset, we have designed 12 open set DA tasks: Ar$\rightarrow$ Cl, Ar$\rightarrow$ Pr, Ar$\rightarrow$ Rw, Cl$\rightarrow$ Ar, Cl$\rightarrow$ Pr, Cl$\rightarrow$ Rw, Pr$\rightarrow$ Ar, Pr$\rightarrow$ Cl, Pr$\rightarrow$ Rw, Rw$\rightarrow$ Ar, Rw$\rightarrow$ Cl and Rw$\rightarrow$ Pr.
	\textbf{ImageNet-Caltech} is a combination of ImageNet-1K \cite{russakovsky2015imagenet} consisting 1000 categories and Caltech-256 with 256 categories. In line with previous works \cite{cao2019learning, cao2018partial}, we have used the common 84 classes as the known or shared classes $C$ and have used the remaining classes as the 'unknown' class in the target domain. We have performed two open set DA tasks I$\rightarrow$ C and C$\rightarrow$ I for this dataset.
	
	\subsection{Evaluation Details} 
	\textbf{Evaluation Protocols.} In this paper, we have followed the evaluation protocol of the Visual Domain Adaptation (VisDA 2018) Open-Set Classification Challenge. This protocol assumes all the target domain private classes $|\overline{C_t}|$ as a unified `unknown' class and the average per-class accuracy for all the $|C|+1$ classes is the final result. Also, being inspired by \cite{saenko2010adapting}, we present the normalized classification accuracy measured on all the known classes and the `unknown' classes ($|C|+1$) as OS, and the normalized classification accuracy only on the shared classes ($|C|$) as OS$^\star$.
	
	\textbf{Implementation Details.} We have used ImageNet pre-trained ResNet-50 and ResNet-152 \cite{he2016deep} with new fully-connected and batch normalization layers as the feature generator. We have used SGD with a learning rate of $0.001$ for pre-trained layers, 10 times higher than that for new layers, and momentum of $0.9$. Note, while the original papers show results on a variety of backbone networks such as VGG, AlexNet, and ResNet-50, for the sake of fairness and consistency we tested them all using ResNet-50. Also, we have executed up to 1000 epochs for training, but if any of the contemporary methods converged earlier than that, we stopped the training. The performance difference between our implementation of the contemporary DA methods and the original papers is mainly due to different backbone networks and the number of iterations. Note that the results cannot be directly compared against other publicly reported results due to different train-test data split and versions of PyTorch.
	
	\textbf{Compared Domain Adaptation Methods.} For the sake of thoroughness, we have compared the performance of the proposed method with: \textbf{1)} Classifier without DA: ResNet \cite{he2016deep} (Note that Negative transfer is calculated against this non-DA classifier.); \textbf{2)} Closed-set DA methods: Domain-Adversarial Neural Networks (DANN) \cite{ganin2016domain} and Residual Transfer Networks (RTN) \cite{long2016unsupervised}; \textbf{3)} Partial DA methods: Importance Weighted Adversarial Nets (IWAN) \cite{zhang2018importance} and Example Transfer Network (ETN) \cite{cao2019learning}; \textbf{4)} Open set DA methods: Unsupervised domain adaptation by back-propagation (BP) \cite{ganin2015unsupervised} with unknown source samples, Assign-and-Transform-Iteratively (ATI) \cite{panareda2017open}, Open Set domain adaptation by Back-Propagation (OSBP) \cite{saito2018open} and  Separate  to  Adapt  (STA) \cite{liu2019separate}; \textbf{5)} Universal DA method: Universal Adaptation Network (UAN) \cite{You2019UniversalDA}. 
	
	\subsection{Classification results}
	\label{results}
	The classification results on the 12 tasks of Office-Home, 6 tasks of Office-31, the task of VisDA and 2 large-scale tasks of ImageNet-Caltech are shown in Tables \ref{t1}, \ref{t2}, \ref{t3} and \ref{t4} respectively (‘-’ indicates that results could not be regenerated because of closed set DA setting and negative transfer for DA classifiers against the non-DA classifier ResNet \cite{he2016deep} (shown in gray background) are indicated by showing the classification accuracy in italic). Our proposed method outperforms all the compared methods in terms of the average per-class accuracy except for Office-31 dataset, where STA \cite{liu2019separate} leads by 0.1\%. However, we have better OS$^\star$ than STA for Office-31 dataset. We observe that all contemporary partial, universal, and open set DA methods lag behind ResNet classifier \cite{he2016deep} on some tasks because of negative transfer during adaptation. This negative transfer is the effect of the difference in source and target domain label space introduced by unknown classes. In addition, the accuracy of the OS$^\star$ is lower than that of OS for the majority of the tasks, which means a large number of unknown images are misclassified. Our proposed method yields better results for OS$^\star$ than OS which indicates a better separation of known and unknown target samples.
	
	DA methods such as BP (with unknown classes in the source domain) \cite{ganin2015unsupervised} and ATI \cite{panareda2017open} are trained to align target domain towards source domain employing distribution matching methods. During this process, the unknown source or target samples disturb the known class feature alignment leading to such performance degradation. The performance lag in OSBP \cite{saito2018open} method compared to ResNet backbone for some tasks supports our claim. That is, the OSBP incurs negative transfer because it tends to align some unknown target samples to known classes by enforcing empirical boundary threshold, and not exploring underlying domain information of target samples. However, because of adapting sample-level transferability, IWAN \cite{zhang2018importance}, ETN \cite{cao2019learning}, UAN \cite{You2019UniversalDA}, and STA \cite{liu2019separate} have lower negative transfer rate compared to other existing methods.  
	
	\begin{figure}[!h]
		\centering
		\begin{subfigure}{.5\columnwidth}
			\centering
			\includegraphics[width=2.5cm,height=1.5cm]{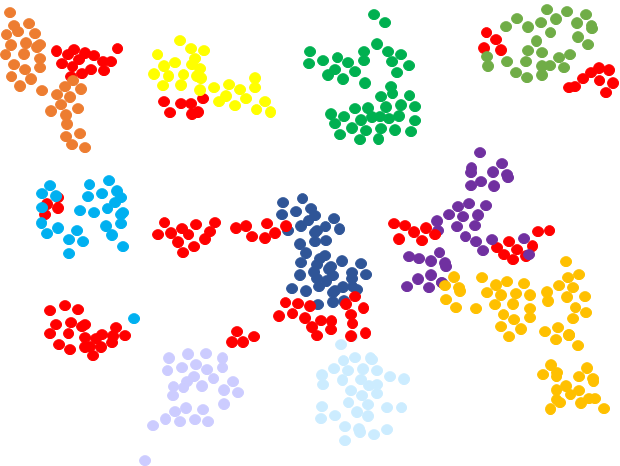}
			\caption{RTN}
			\label{ts1}
		\end{subfigure}%
		\begin{subfigure}{.5\columnwidth}
			\centering
			\includegraphics[width=2.5cm,height=1.5cm]{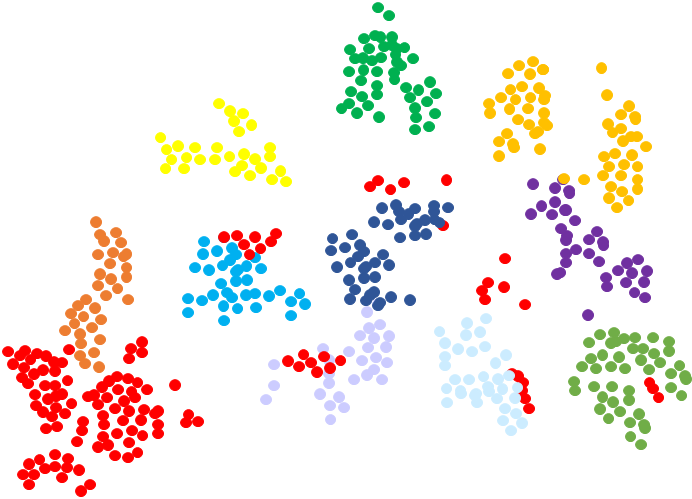}
			\caption{ATI}
			\label{ts2}
		\end{subfigure}
		\begin{subfigure}{.5\columnwidth}
			\centering
			\includegraphics[width=2.5cm,height=1.5cm]{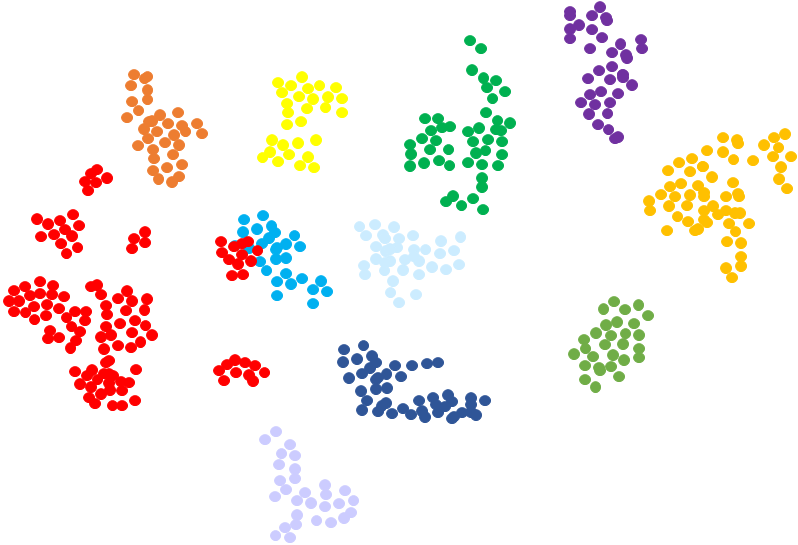}
			\caption{ETN}
			\label{ts3}
		\end{subfigure}%
		\begin{subfigure}{.5\columnwidth}
			\centering
			\includegraphics[width=2.5cm,height=1.5cm]{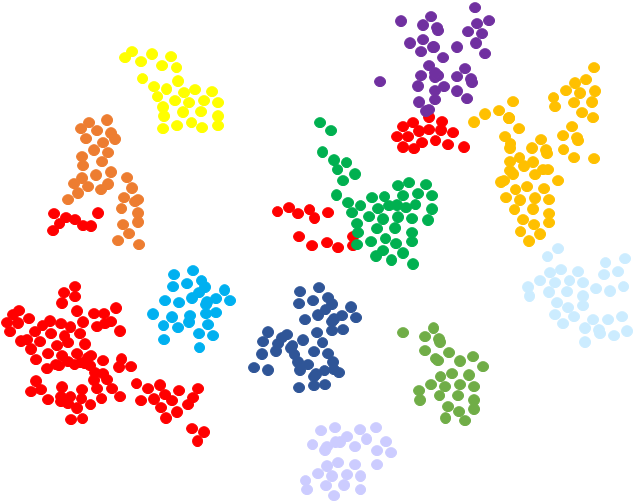}
			\caption{IWAN}
			\label{ts4}
		\end{subfigure}
		\begin{subfigure}{.5\columnwidth}
			\centering
			\includegraphics[width=2.5cm,height=1.5cm]{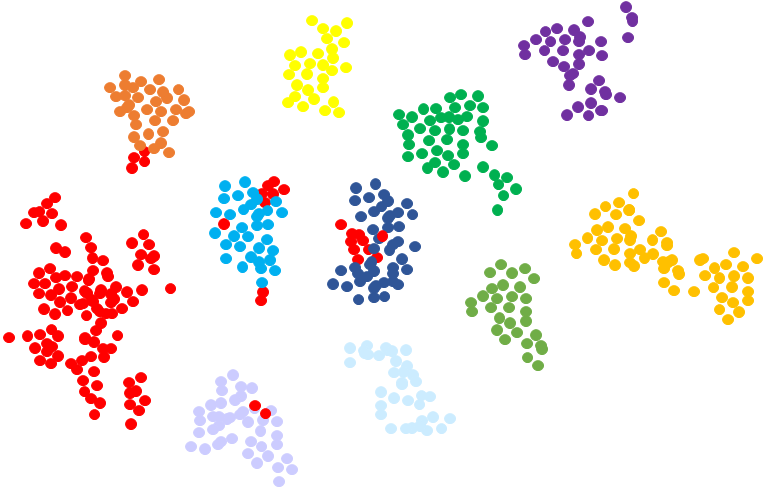}
			\caption{OSBP}
			\label{ts5}
		\end{subfigure}%
		\begin{subfigure}{.5\columnwidth}
			\centering
			\includegraphics[width=2.5cm,height=1.5cm]{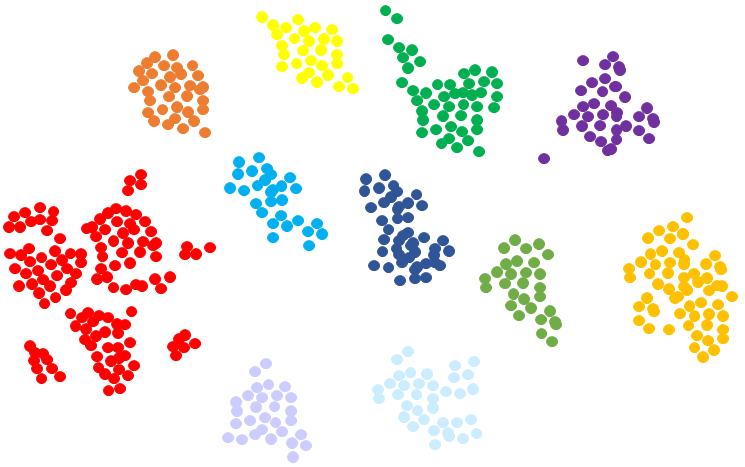}
			\caption{Our}
			\label{ts6}
		\end{subfigure}
		\caption{Learned features of our proposed method for \textbf{A }$\rightarrow$ \textbf{W} task from \textbf{Office-31} dataset show a better separation of unknown samples (red dots) from known classes than contemporary DA methods. We select 10 shared classes, 10 source domain private classes, and 10 target domain private classes for the task. Visualization prepared by the t-SNE algorithm \cite{maaten2008visualizing} with the Perplexity parameter set to 50.}
		\label{tsne}
	\end{figure}
	
	In Fig. \ref{tsne}, we plot the t-SNE \cite{maaten2008visualizing} embeddings of the features learned by RTN \cite{long2016unsupervised}, ATI \cite{panareda2017open}, IWAN \cite{zhang2018importance}, ETN \cite{cao2019learning}, OSBP \cite{saito2018open} and proposed method on \textbf{A }$\rightarrow$ \textbf{W} task with 10 shared classes, 10 source domain private classes and 10 target domain private classes as per respective DA settings. 
	The proposed method demonstrates significantly comprehensible and well-segregated clusters for all known and unknown classes than other DA methods. This distinct separation of unknown target samples from the known ones is because of the supplementary source classifier, which is trained only on the source samples to learn discriminative features for known classes. Unlike OSBP \cite{saito2018open}, we do not need any prior training on the
	\begin{table*}[!ht]
		\caption{Accuracy (\%) of proposed method on VisDA2017, Office-31, and ImageNet-Caltech tasks for ablation study.}
		\vspace{-4.5mm}
		\label{t5}
		\begin{center}
			\resizebox{\textwidth}{!}{%
				\begin{tabular}{cccccccccccc}
					\hline
					\multirow{3}{*}{Approach} & \multicolumn{11}{c}{Accuracy (\%)}\\\cline{2-12}
					& VisDA & A$\rightarrow$ W & D$\rightarrow$ W & W$\rightarrow$ D & A$\rightarrow$ D & D$\rightarrow$ A & W$\rightarrow$ A & Avg & I$\rightarrow$ C & C$\rightarrow$ I&Avg\\ \hline
					w/o $d_2$ &63.3&86.1 &93.1 &96.2 &84.2 &86.5 &82.1 & 88.0& 75.3& 66.4&70.9\\
					w/o $d_1$&61.1&84.8 &93.8 &94.7 &85.1 &84.7 &82.3 & 87.6& 74.6& 65.0&69.8\\
					Our(ResNet-50) &65.9&88.3 &97.3 &98.1 &87.8 &89.9 &84.9 & 91.1& 77.4& 69.8&73.6\\\hline
				\end{tabular}
			}
		\end{center}
	\end{table*}
	source domain to learn discriminative known class features to support better classification. On the other hand, RTN \cite{long2016unsupervised} and ATI \cite{panareda2017open} methods which utilize distribution matching techniques such as MMD only tries to align target samples with the source ones and do not separate well among known and unknown classes. Though ETN \cite{cao2019learning} shows better clusters than IWAN \cite{zhang2018importance} for initiating less negative transfers, it certainly lags behind our method.
	
	\begin{figure}[!h]
		\centering
		\includegraphics[width=.65\columnwidth,height=4.7cm]{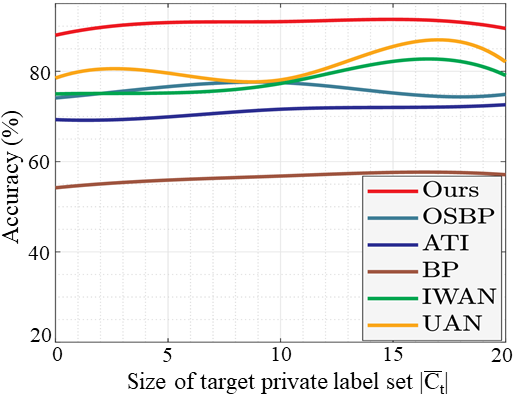}
		\caption{The proposed method consistently performs better than contemporary domain adaptation methods for all cases of $|\overline{C_t}|$. The performance of our proposed method increases with an increase of the number of unknown classes in the target domain $|\overline{C_t}|$. 
		}
		\label{a1}
	\end{figure}
	\begin{figure}[!h]
		\centering
		\includegraphics[width=.65\columnwidth,height=4.5cm]{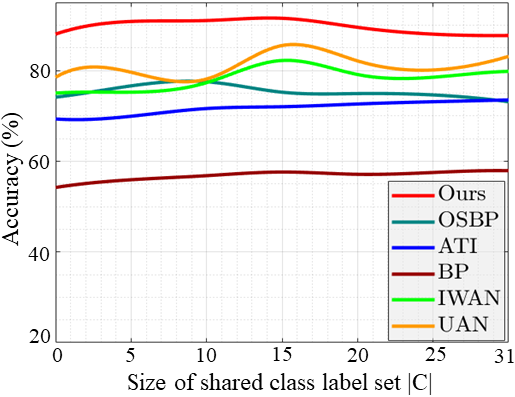}
		\caption{Our proposed method outperforms contemporary domain adaptation methods with different domain adaptation settings for different sizes of the shared label set. We observe that when $|C|$ reaches beyond 20, the performance of our method decreases slightly. This indicates our proposed method is more suitable for tasks that have more `unknown' classes in the target domain.}
		\label{a2}
	\end{figure}
	
	\subsection{Analysis on Various Open Set DA Settings}
	\textbf{Varying Size of $\overline{\textit{C\textsubscript{t}}}$.} We compare the performance of our proposed method with other methods by varying the number of unknown samples in the target domain for D$\rightarrow$ A task. Fig. \ref{a1} shows that our proposed method maintains moderate increment of performance with no significant drop in-between transitions of varying $|\overline{C_t}|$ and outperforms all compared DA methods for all cases. This indicates a larger number of unknown classes in the target domain $|\overline{C_t}|$ compared to the shared classes assists both the source $G_{C_1}$ and supplementary source $G_{C_1}$ classifiers, by initiating less distraction and can lead to solving more realistic tasks where unknown source samples are unavailable. Note that our proposed method does not take advantage of any prior knowledge about the label sets like OSBP \cite{saito2018open} and IWAN \cite{zhang2018importance}.
	
	\textbf{Varying Size of \textit{C}.} We further explore the performance of the proposed method by varying the number of shared classes for the same task D$\rightarrow$ A and compare it with other methods (Fig. \ref{a2}). The proposed method maintains high accuracy with a slight drop when $|C| > 20$. The evaluation of the task shown here is a sub-task of the Office-31 dataset. The office-31 dataset has 31 categories when the shared label set $C$ is beyond 20 the target private label set $|\overline{C_t}|$ decreases to less than 10. The proposed method is designed to handle well the tasks which have a large number of unknown classes in the target private label space. Therefore, the increment of shared label space, which causes decrement of target private label space in a large proportion, harms the model. However, the proposed method substantially outperforms other compared methods for all cases of $|C|$.
	\subsection{Ablation Study}
	We execute an ablation study for evaluating two variants of the proposed domain adaptation method with the multi-classifier based weighting module to investigate deeper into its effectiveness. \textbf{w/o d\textsubscript{2}} is the variant of proposed method without integrating the outcome of $d_2(x)$ (domain information) in the procedure of weighting target samples, i.e. in Equations (\ref{e8}) and (\ref{e9}). \textbf{w/o d\textsubscript{1}} is the variant without integrating the known-source label information into the weighting mechanism by omitting $G_{C_2}$ and deploying $G_D$ to depend only on the value of $d_2(x)$ for source and target samples. To execute this variant, we need to omit Equations (\ref{e6}) and (\ref{e7}), omit $d_1(x)$ in Equations (\ref{e8}) and (\ref{e9}).
	
	Table \ref{t5} presents the results for the variants of the proposed method, as mentioned above. The performance of both \textbf{w/o d\textsubscript{2}} and \textbf{w/o d\textsubscript{1}} lag behind the proposed method. This is because both the deciding factors $d_1(x_i^t)$ and $d_2(x_i^t)$ are required for defining meaningful weights $W(x_j^t)$ to the target samples. We also observe that the variant without \textbf{w/o d\textsubscript{2}} achieves better average per-class accuracy than the other one for all three tasks, which indicates integrating the supplementary source classifier $G_{C_2}$ in the weighting module for exploiting label information is more effective.
	\subsection{Weight Visualization}\label{weights}
	\begin{figure}[!ht]
		\centering
		\begin{subfigure}[b]{.49\columnwidth}
			\centering
			\includegraphics[width=\textwidth]{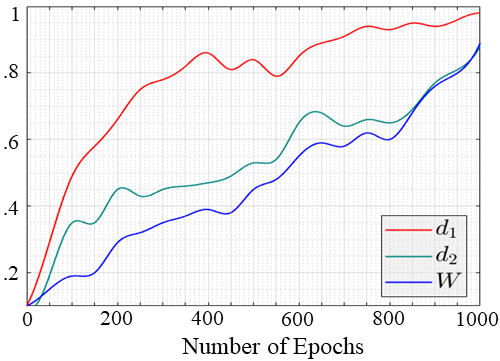}
			\caption[]%
			{{\small }}    
			\label{6a}
		\end{subfigure}
		\hfill
		\begin{subfigure}[b]{.49\columnwidth}  
			\centering 
			\includegraphics[width=\textwidth]{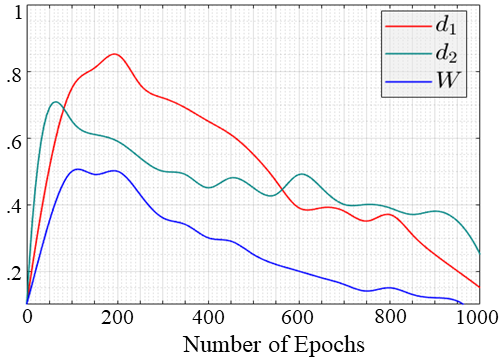}
			\caption[]%
			{{\small }}    
			\label{6b}
		\end{subfigure}
		\caption[ ]
		{Pictorial representation of learned weights of (a) known target samples and (b) unknown target samples for the task D$\rightarrow$ A (Office-31). Here, $W$ represents the weights assigned to target samples during training by $G_D$ after assessing the two similarity measures $d_1$ and $d_2$.}
		\label{6}
	\end{figure}
	To analyse the trend of generated weights, we plot the learned similarity measures ($d_1$ and $d_2$) and final weights ($W$) of known and unknown target samples against the training epochs in Figure \ref{6}. Figures \ref{6a} and \ref{6b} show that our proposed method assigns sufficiently high and low weights ($W$) to known and unknown target samples, respectively. It is evident that such weights will assist the adversarial module in enhancing positive transfer by better separation of known target samples from unknowns. 
	During the early stage of learning, both $d_1$ and $d_2$ increases for known and unknown samples which means the classifiers give uncertain decisions. However, after some initial epochs, for known samples, $d_1$ seems to increase at a larger pace and learns to yield a much higher value than $d_2$. This is because the classifier $G_{C_2}$ is trained 
	only on source samples. On the other hand, for unknown samples, both the similarity measures start to decrease. When both $G_{C_2}$ and $G_{C_1}$ converge to their optimal value for $G_F$, $d_1$, $d_2$ and $W$ for known target samples reaches near $0.98$, $0.88$ and $0.86$ respectively. Which means the target samples get very high weights as the training proceeds and dictate $G_F$ to decrease the ‘unknown’ class probability for aligning them to known classes. For unknown target samples, $d_1$, $d_2$ and $W$ decreases to $0.15$, $0.25$ and $0.04$ respectively. This indicates the target samples get very low weights and assists $G_F$ to increase the ‘unknown’ class probability and align them to unknown classes. It is worth mentioning, the final value of $d_1$, $d_2$ and hence $W$ may differ a little based on the dataset or task.
	
	\section{Conclusion}
	In this paper, we propose a multi-classifier based domain adversarial network for an open set domain adaptation setting, where the target domain has a larger number of classes than the source domain. 
	In our proposed method, the multi-classifier structure poses a weighting module that explores discriminative label and domain information, and assign distinguishable scores to the known and unknown target samples for enhancing positive transfer and eventually, assisting the feature generator and the domain classifier to separate known and unknown target samples. Another noteworthy attribute of our proposed method is the ability to discover the boundary between shared label space and target private label space automatically. A thorough experimental evaluation has demonstrated that the proposed method consistently outperforms the existing DA methods. We have further shown through the ablation study that the two deciding factors for generating weights for target samples are crucial for maintaining the integrity of the model and initiating positive sample transfer.
	
	

	
	
	%

	



	\ifCLASSOPTIONcaptionsoff
	\newpage
	\fi

	
	
	%

	\bibliographystyle{IEEEtran}
	\bibliography{ref}
	
	
	%
	\vspace{-12mm}
	\begin{IEEEbiography}[{\includegraphics[width=1in,height=1.25in,clip,keepaspectratio]{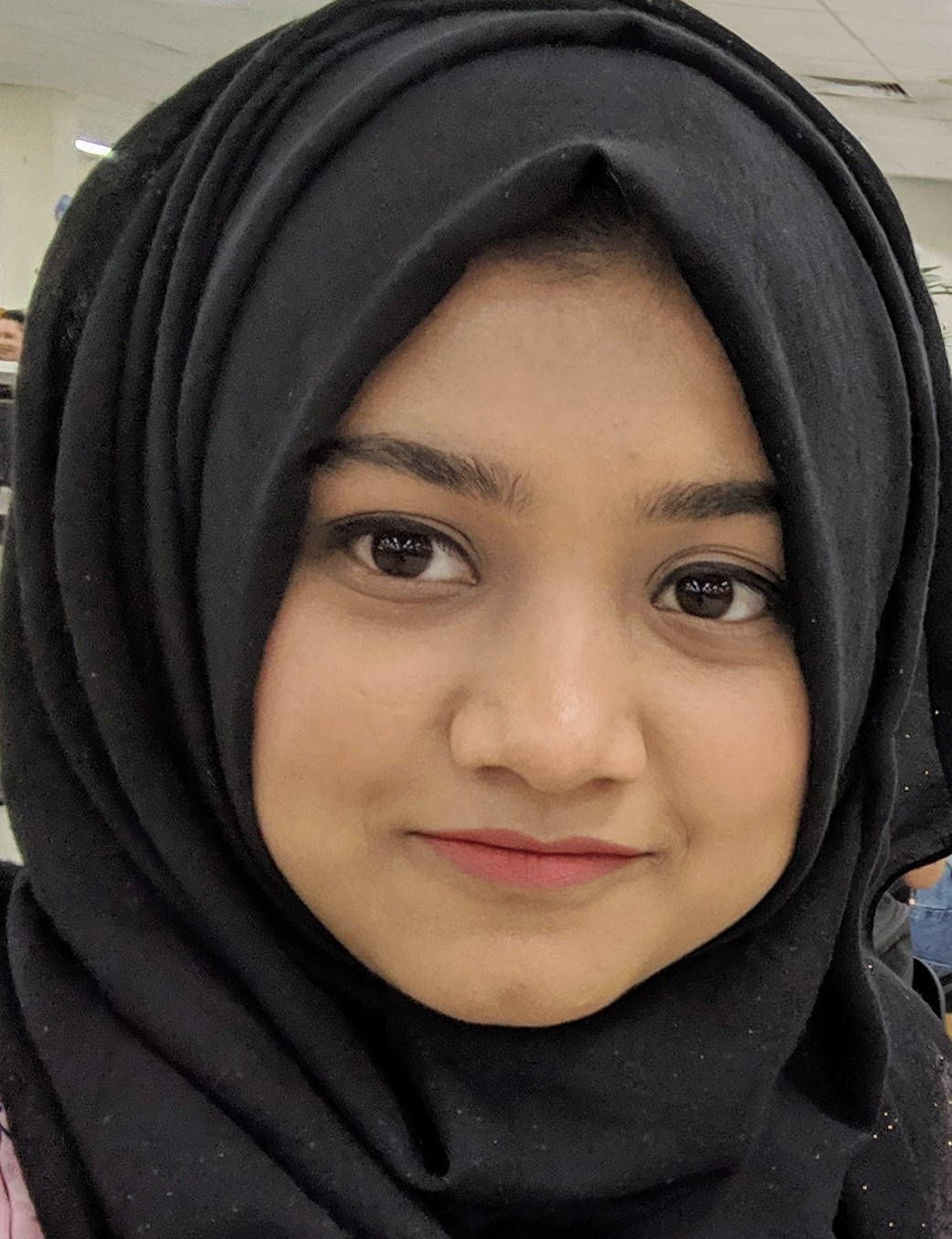}}]{Tasfia Shermin} has received a bachelor’s degree in Computer Science and Engineering (CSE) from the Military Institute of Science and Technology, Bangladesh, in 2018.  She is currently pursuing the Doctor of Philosophy (Ph.D.) degree at the School of Science, Engineering and Information Technology, Federation University, Australia. Her research interests include artificial intelligence in computer vision, deep learning and transfer learning. 
	\end{IEEEbiography}
	\vspace{-12mm}
	\begin{IEEEbiography}[{\includegraphics[width=1in,height=1.25in,clip,keepaspectratio]{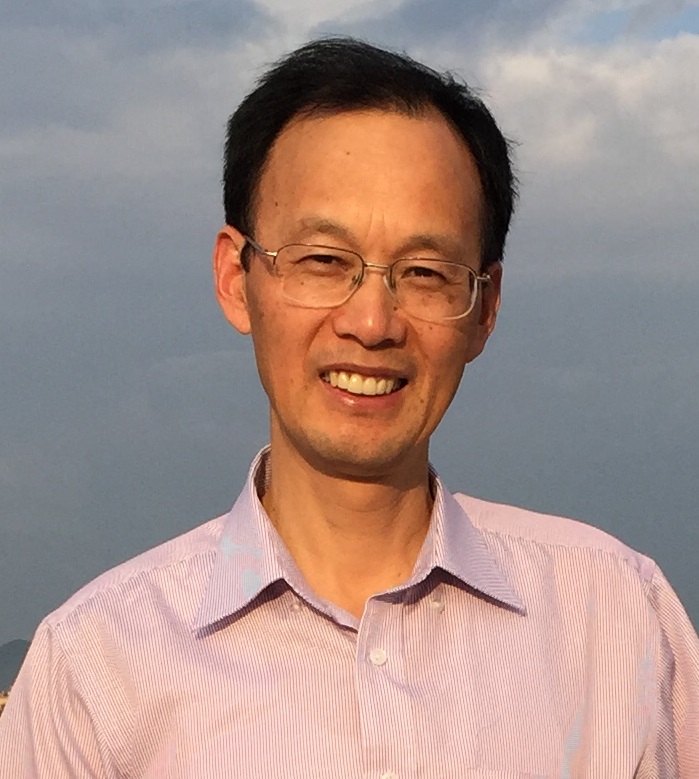}}]{Guojun Lu}  is a Professor in the School of Science, Engineering and Information Technology, Federation University Australia. He has many years’ research experience in artificial intelligence, multimedia signal processing and retrieval, and has successfully supervised over 20 PhD students. He has held positions at Loughborough University, National University of Singapore, Deakin University and Monash University, after he obtained his PhD from Loughborough University and BEng from Nanjing Institute of Technology (now South East University, China). He has published over 230 refereed journal and conference papers and wrote two books.
	\end{IEEEbiography}
	\vspace{-10mm}
	\begin{IEEEbiography}[{\includegraphics[width=1in,height=1.25in,clip,keepaspectratio]{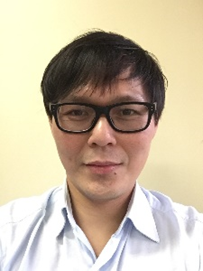}}]{Shyh Wei Teng}  is an Associate Professor and Deputy Dean at School of Science, Engineering and IT, Federation University Australia. His research interests include Image/video processing; Machine learning; and Multimedia analytics. He has so far published over 70 refereed research papers. He received various competitive research funding at state, federal and international levels.
	\end{IEEEbiography}
	\vspace{-10mm}
	\begin{IEEEbiography}[{\includegraphics[width=1in,height=1.25in,clip,keepaspectratio]{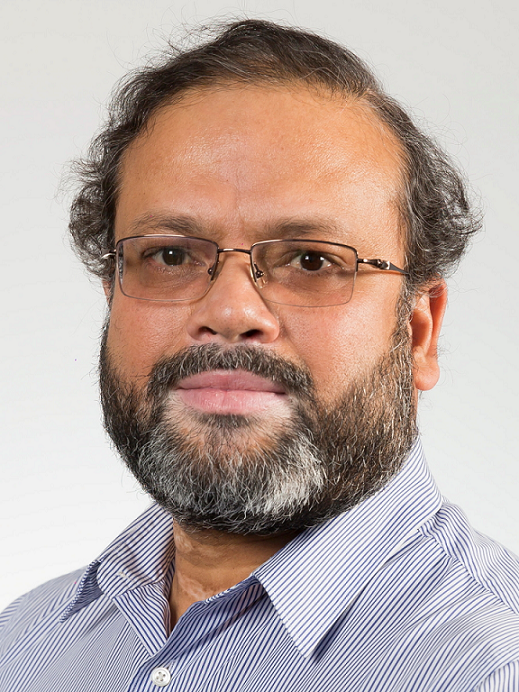}}]{Manzur Murshed} (M’96–SM’12) received a BScEngg (Hons) degree in computer science and engineering from Bangladesh University of Engineering and Technology in 1994 and a PhD degree in computer science from the Australian National University in 1999. Currently, he is a Professor and the Associate Dean (Research) at the School of Science, Engineering and Information Technology, Federation University Australia. His research interests include video technology, machine learning, wireless communications, Cloud computing, and security privacy. He has published 230+ refereed research papers, received \$5m research grants, and supervised 25 PhDs. He is serving IEEE TMM and served IEEE TCSVT as an Associate Editor.
	\end{IEEEbiography}
	\vspace{-12mm}
	\begin{IEEEbiography}[{\includegraphics[width=1in,height=1.25in,clip,keepaspectratio]{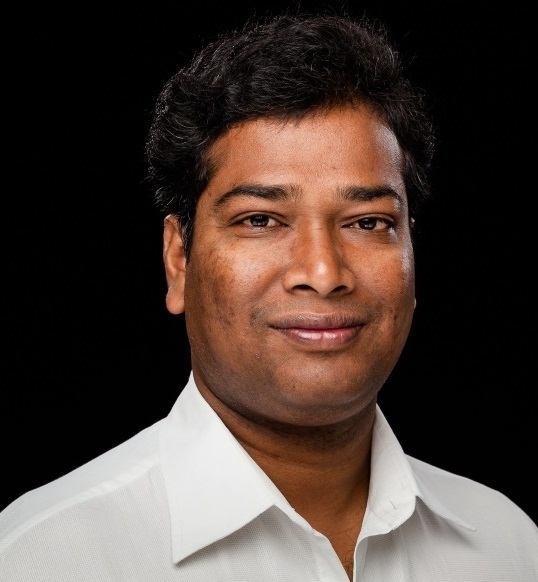}}]{Ferdous Sohel} (M’08–SM’13) received his PhD degree from Monash University, Australia, in 2009. He is currently an Associate Professor in Information Technology at Murdoch University, Australia. Prior to his joining Murdoch University, he was a Research Assistant Professor at the University of Western Australia from 2008 to 2015. His research interests include computer vision, image processing, machine learning, digital agritech, health analytics, cyber forensics, and video coding. He has been serving as an Associate Editor of IEEE Transactions on Multimedia.
	\end{IEEEbiography}
	


	
	

\end{document}